\documentclass[10pt,twocolumn,letterpaper]{article}

\usepackage{iccv}
\usepackage{times}
\usepackage{epsfig}
\usepackage{graphicx}
\usepackage{amsmath}
\usepackage{amssymb}
\usepackage{pifont}
\usepackage{colortbl}
\usepackage{multirow}
\usepackage{diagbox}
\usepackage{color,xcolor}
\definecolor{fired}{RGB}{222,82,57}
\definecolor{iceblue}{RGB}{33,102,200}
\usepackage[export]{adjustbox}
\definecolor{mygray}{gray}{.9}

\makeatletter
\newcommand{\thickhline}{%
    \noalign {\ifnum 0=`}\fi \hrule height 1pt
    \futurelet \reserved@a \@xhline
}

\usepackage[pagebackref=true,breaklinks=true,letterpaper=true,colorlinks,bookmarks=false]{hyperref}

\iccvfinalcopy 

\def\iccvPaperID{6523} 

\ificcvfinal\fi

\begin{document}

\title{E$^2$VPT: An Effective and Efficient Approach for Visual Prompt Tuning 
}
\author{$\text{Cheng Han}^{1}$, $\text{Qifan Wang}^{2}$, $\text{Yiming Cui}^{3}$, $\text{Zhiwen Cao}^{4}$, $\text{Wenguan Wang}^{5}$, $\text{Siyuan Qi}^{6}$, $\text{Dongfang Liu}^{1}$\thanks{Corresponding author} \\
$\text{Rochester Institute of Technology}^{1}$, $\text{Meta AI}^{2}$, $\text{University of Florida}^{3}$, $\text{Purdue University}^{4}$ \\
$\text{Zhejiang University}^{5}$, $\text{BIGAI}^{6}$\thanks{National Key Laboratory of General Artificial Intelligence, Beijing Institute for General Artificial Intelligence} \\
{{\tt\small \{ch7858,~dongfang.liu\}@rit.edu, wqfcr@fb.com, cuiyiming@ufl.edu, cao270@purdue.edu}} \\
{{\tt\small wenguanwang.ai@gmail.com, syqi@bigai.ai}}
}

\maketitle
\ificcvfinal\thispagestyle{empty}\fi

\begin{abstract}
As the size of transformer-based models continues to grow, fine-tuning these large-scale pretrained vision models for new tasks has become increasingly parameter-intensive.
Parameter-efficient learning has been developed to reduce the number of tunable parameters during fine-tuning. Although these methods show promising results, there is still a significant performance gap compared to full fine-tuning.
To address this challenge, 
we propose an Effective and Efficient Visual Prompt Tuning (E$^2$VPT) approach for large-scale transformer-based model adaptation. Specifically, we introduce a set of learnable key-value prompts and visual prompts into self-attention and input layers, respectively, to improve the effectiveness of model fine-tuning. Moreover, we design a prompt pruning procedure to systematically prune low importance prompts while preserving model performance, which largely enhances the model's efficiency.
Empirical results demonstrate that our approach outperforms several state-of-the-art baselines on two benchmarks, with considerably low parameter usage (\eg, 0.32\% of model parameters on VTAB-1k). Our code is available at \href{https://github.com/ChengHan111/E2VPT}{https://github.com/ChengHan111/E2VPT}.

\end{abstract}

\vspace{-1em}
\section{Introduction}
\label{sec:introduction}
The development of artificial intelligence (AI) should not only prioritize performance advances, but also emphasize sustainable deployment~\cite{nishant2020artificial,van2021sustainable,vinuesa2020role,wu2022sustainable}. Despite the captivating pursuit of performance improvements in visual-related tasks, the size of present models has been rapidly increasing, resulting in energy-intensive and computationally expensive training~\cite{innes2019differentiable,sanh2019distilbert,you2019large}. 
Transformer-based architectures currently dominate visual-related models, such as ViT-Huge~\cite{dosovitskiy2020image} (632M) and Swin-Large~\cite{liu2021swin} (197M), with significantly more parameters than the Convolutional Neural Networks (CNN) like ResNet~\cite{he2016deep} (25M). 
Training such large models from scratch presents challenges such as limited data~\cite{brown2020language,guo2020parameter,tajbakhsh2016convolutional} and slow convergence at low accuracy~\cite{kaiser2017one,lin2021traceability}. 
A common paradigm to overcome these challenges is \textit{pretrain-then-finetune}, which reduces the need for vast amounts of training data and speeds up processing of various visual tasks. However, the traditional \textit{full fine-tuning} involves storing and deploying a complete copy of the backbone parameters for every single task~\cite{jia2022visual}, which remains computationally expensive and not suitable for fast model deployment.

\begin{figure}[t!]
  \centering
\includegraphics[width=0.50\textwidth]{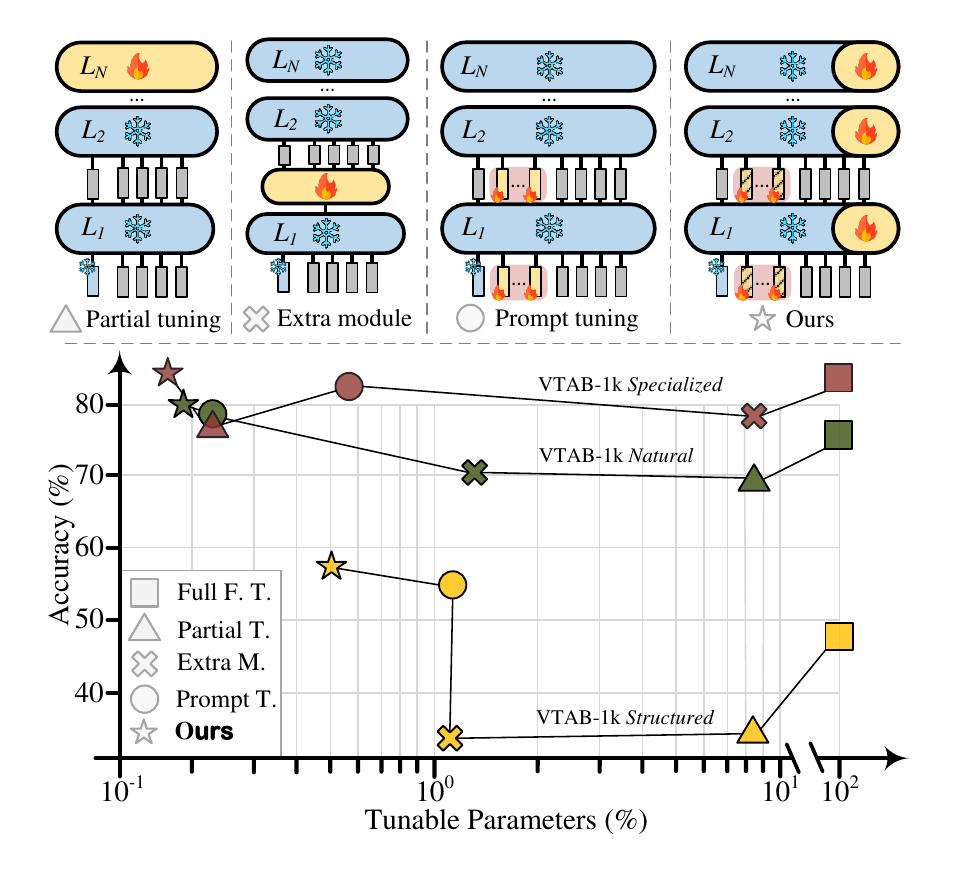}
\vspace{-15pt}
\caption{\textbf{E$^{2}$VPT (ours) $vs$ concurrent arts} (\ie, \protect\includegraphics[scale=0.05,valign=c]{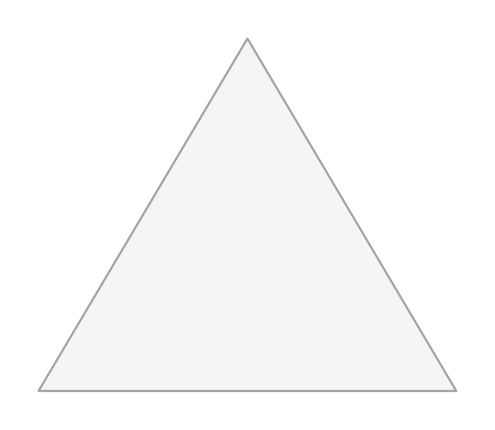} partial tuning \cite{yosinski2014transferable}, \protect\includegraphics[scale=0.05,valign=c]{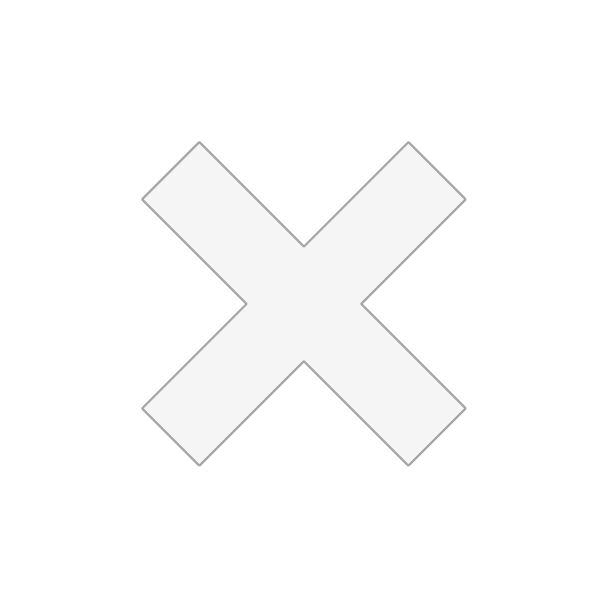} extra module \cite{cai2020tinytl}, and \protect\includegraphics[scale=0.05,valign=c]{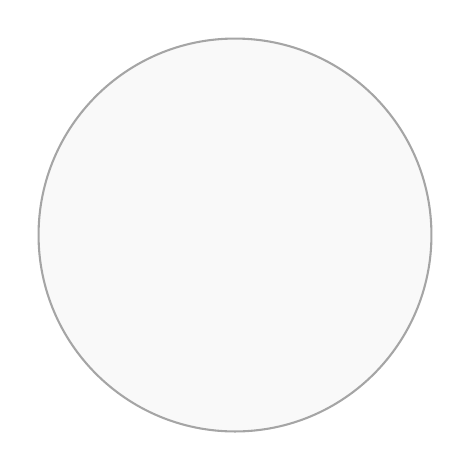} prompt tuning \cite{jia2022visual} methods) under \textit{pretrain-then-finetune} paradigm. Our method yields solid performance gains over state-of-the-art fine-tuning methods and competitive to full fine-tuning on a wide range of classification tasks adapting the pretrained ViT-Base/16~\cite{dosovitskiy2020image} as backbone with considerable lower parameter usage (see Table~\ref{table:fgvc_vtab_main}). \protect\includegraphics[scale=0.10,valign=c]{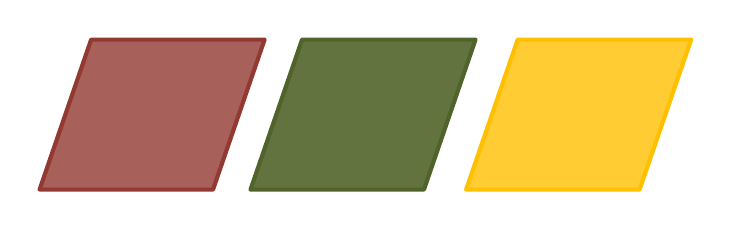} colors represent results on VTAB-1k~\cite{zhai2019large} \textit{Specialized}, \textit{Natural} and \textit{Structure}, respectively.}
\label{fig:plot1}
\vspace{-18pt}
\end{figure}

To address this issue, various approaches have been developed, which can be divided into three main categories (see Fig.~\ref{fig:plot1}): partial tuning, extra module, and prompt tuning methods.
\textit{Partial tuning} methods~\cite{chen2021empirical,jia2021exploring,mahajan2018exploring} only fine-tune part of the backbone, such as the classifier head or last few layers, while freezing the others. \textit{Extra module} methods insert learnable bias term~\cite{cai2020tinytl} or additional adapters~\cite{rebuffi2017learning, zhang2020side} to the network for adaptation. \textit{Prompt tuning} methods add prompt tokens~\cite{jia2022visual, ju2022prompting, zang2022unified} to the input layer of the transformer without changing or fine-tuning the backbone itself.
All of these methods operate within the \textit{pretrain-then-finetune} paradigm, which reduces the number of learnable parameters compared to full fine-tuning~\cite{chen2021empirical,jia2021exploring,mahajan2018exploring,rebuffi2017learning, zhang2020side}. 
However, despite achieving promising results, there are two main limitations in existing parameter-efficient methods.
\textbf{Firstly}, they do not scrutinize the core architecture of the transformer's self-attention mechanism, resulting in a large performance gap with full fine-tuning. \textbf{Secondly}, they usually need to fine-tune a relatively large number of parameters to achieve reasonable performance and fail to explore the extremes of parameter efficiency.



The perspective outlined above leads to two fundamental questions: \textit{\textbf{\ding{182}} How can we establish the \textbf{effectiveness} of prompt tuning for large-scale transformer-based vision models? \textbf{\ding{183}} How can we explore the extremes of parameter \textbf{efficiency} to reduce the number of tunable parameters?}
These two questions are the foundation of our work. 
The intuition is that instead of solely focusing on modifying inputs, as in previous prompt tuning methods, we should explicitly investigate the potential of improving the self-attention mechanism during fine-tuning, and explore the extremes of parameter efficiency.

In response to question \textbf{\ding{182}}, we discuss and analyze the self-attention mechanism of the transformer, which is crucial in capturing long-range token dependencies within a global context~\cite{han2022survey,khan2022transformers,lin2022survey}. In additional to the input visual prompts, we introduce learnable key-value prompts and integrate them into the Key and Value matrices in the self-attention layers. The key-value prompts are jointly learned with the input visual prompts during fine-tuning. This approach effectively leverages the well-designed prompt architecture of the transformer, resulting in significant performance improvements. Moreover, it provides a generic plug-and-play prompt module for current transformer architectures, and its fine-tuning solution is conceptually different from all aforementioned arts in the vision domain.

Motivated by \textbf{\ding{183}}, we propose a pruning strategy to further reduce the number of parameters while maintaining the model performance. 
Our approach draws inspiration from the lottery ticket hypothesis (LTH)~\cite{frankle2018lottery, zhuang2023survey}, which posits that for a given task, there exists a sub-network that can match the test accuracy of the original over-parameterized network without the unnecessary weights~\cite{han2015learning,hassibi1992second,lecun1989optimal,li2022automated,li2016pruning}. Building on this paradigm, we revisit the core design of prompt tuning methods and further reduce the number of learnable parameters. Specifically, we aim to retain the prompt tokens that contribute significantly to the performance, while pruning the prompt tokens that are redundant or unnecessary during fine-tuning. By pruning these unnecessary prompts, we can significantly improve the prompt tuning efficiency while maintaining the performance.


To answer question \textbf{\ding{182}}-\textbf{\ding{183}}, we propose \textbf{E$^{2}$VPT}, namely \textbf{E}ffective and \textbf{E}fficient \textbf{V}isual \textbf{P}rompt \textbf{T}uning. ${\rm E^{2}VPT}$ is a novel prompt tuning framework that is both architecture-aware and pruning-anchored (see Fig.~\ref{fig:plot1}). 
In \S\ref{sec:related_work}, we conduct a literature review and discuss relevant works. Our proposed approach is presented in \S\ref{sec:mixed_prompt}, where we describe in detail how we design visual and key-value prompts to achieve superior performance with fewer parameters.$_{\!}$  In~\S\ref{sec:Experiment}, we$_{\!}$  present$_{\!}$  compelling$_{\!}$  experimental$_{\!}$  results$_{\!}$  on$_{\!}$  various$_{\!}$  benchmarks,$_{\!}$  backbones,$_{\!}$  and$_{\!}$  different$_{\!}$  pretraining$_{\!}$  objectives.$_{\!}$  Specifically, our$_{\!}$  approach$_{\!}$  achieves$_{\!}$  an$_{\!}$  average$_{\!}$  improvement$_{\!}$  of \textbf{5.85\%} in accuracy on VTAB-1k compared$_{\!}$  to$_{\!}$  full$_{\!}$  fine-tuning, and \textbf{1.99\%} compared to VPT~\cite{jia2022visual}. Moreover,$_{\!}$  our$_{\!}$  approach$_{\!}$  uses$_{\!}$  considerably$_{\!}$  fewer$_{\!}$  learnable$_{\!}$  parameters$_{\!}$  than$_{\!}$  existing$_{\!}$  methods,$_{\!}$  accounting$_{\!}$  for$_{\!}$  an$_{\!}$  average of only \textbf{0.32\%} of the backbone parameters on VTAB-1k, whereas VPT on average requires 0.68\% (see Fig.~\ref{fig:plot1}).$_{\!}$  We$_{\!}$  further$_{\!}$  demonstrate$_{\!}$  and$_{\!}$  explain$_{\!}$  the$_{\!}$  superiority$_{\!}$  of$_{\!}$  our$_{\!}$  approach$_{\!}$  over$_{\!}$  VPT$_{\!}$  with$_{\!}$  hyperbolic$_{\!}$  visualization.$_{\!}$  Finally,$_{\!}$  we$_{\!}$  demonstrate$_{\!}$  the$_{\!}$  strong$_{\!}$  algorithmic$_{\!}$  generalization$_{\!}$  of$_{\!}$  our$_{\!}$  approach$_{\!}$  to$_{\!}$  the$_{\!}$  language$_{\!}$  domain$_{\!}$  in$_{\!}$  the$_{\!}$ Appendix. $_{\!}$We$_{\!}$ trust$_{\!}$ that$_{\!}$ this$_{\!}$ work$_{\!}$ provides$_{\!}$ valuable$_{\!}$ insights$_{\!}$ into$_{\!}$ related$_{\!}$ fields.

\vspace{-0.5em}
\section{Related Work}
\label{sec:related_work}

\vspace{-0.1em}
\subsection{Vision Transformers}
\label{subsec:Vision_transformer}
\vspace{-0.1em}

Inspired by the remarkable success of transformers in natural language processing (NLP)~\cite{brown2020language,devlin2018bert,liu2019roberta, raffel2020exploring,vaswani2017attention, wang2022visual}, researchers have extended the transformer architecture to various supervised vision tasks, including image classification~\cite{dosovitskiy2020image,liu2022swin,liu2021swin, lu2023transflow}, image segmentation~\cite{liang2023clustseg, liu2021sg, strudel2021segmenter,wang2021max, wang2022learning, wang2021end,zheng2021rethinking}, object detection~\cite{beal2020toward,carion2020end,liu2020video,pan20213d,yuan2021temporal,zhu2020deformable} and pose estimation~\cite{huang2020hand, huang2020hot, lin2021end, yang2021transpose}). 
Self-supervised pretraining paradigms~\cite{bao2021beit, chen2021empirical, he2022masked} has also been explored, leading to state-of-the-art results.
transformers dominate in visual-related disciplines due to their superior performance and scalability compared to convolutional neural networks (CNNs)~\cite{he2022parameter, jia2022visual}.  However, the significant computational and parameter overhead required to adapt transformers to various vision tasks cannot be ignored~\cite{fournier2021practical, islam2022recent, zhang2022patchformer}.
For instance, recent transformer-based models such as MViTv2-Large~\cite{li2021improved} (218M), ViT-G~\cite{zhai2022scaling} (1.8B), SwinV2-G~\cite{liu2022swin} (3.0B), and V-MoE~\cite{riquelme2021scaling} (14.7B) incur substantial computational costs. Therefore, we propose ${\rm E^{2}VPT}$, which is designed to reduce the computational cost of transformer-based architectures while maintaining high performance in the \textit{pretrain-then-finetune} paradigm.

\subsection{Parameter-efficient Fine-tuning} 
\label{subsec:prior_arts}
Efficient model training has drawn much attention in the vision community, particularly with the rise of Vision Transformers~\cite{arnab2021vivit, chen2021crossvit,dosovitskiy2020image,liu2021swin, wang2021pyramid}. 
However, despite their effectiveness and widespread use, these models are often too large for practical deployment and adaptation. As a result, the \textit{pretrain-then-finetune} paradigm is commonly employed. While full fine-tuning ensures strong performance, it is an expensive approach that involves updating all network parameters~\cite{he2022parameter, tajbakhsh2016convolutional}. To overcome this challenge, researchers are exploring alternatives that balance parameter-efficiency and robust performance, which can be broadly categorized into three groups: \textit{partial tuning}, \textit{extra module} and \textit{prompt tuning} methods.



\textit{Partial tuning} methods are widely used for parameter-efficient fine-tuning. These methods involve freezing most of the backbone and only fine-tune a small portion of the parameters, such as linear~\cite{iofinova2022well} or MLP heads~\cite{chen2020improved}, or a few blocks/layers of the backbone~\cite{he2022masked, noroozi2016unsupervised, yosinski2014transferable, zhang2016colorful}.
While these methods are straightforward and simple to implement~\cite{chen2021empirical, jia2021exploring, mahajan2018exploring}, they often have a large performance gap compared to full fine-tuning. 
\textit{Extra module} methods design additional learnable plug-in architecture for fine-tuning. For example, the work in \cite{zhang2020side} introduces a side structure alternatively while freezing the original network. The works in~\cite{cai2020tinytl, rebuffi2017learning} insert additional residual units into the backbone.
However, one drawback of these methods is that the inserted modules are often customized for specific architectures and might not be generalized to others. Additionally, these modules usually consume even more parameters compared to partial tuning methods.
\textit{Prompt tuning} or prompting~\cite{he2022hyperprompt, lester2021power,ma2022xprompt, yang2023mixpave} has been originally proposed for fast model adaptation in the language domain. These methods prepend a set of learnable vectors to the input of the backbone and only update these task-specific prompts during fine-tuning.
Recently, visual-related prompting~\cite{gao2022visual, jia2022visual, xing2022class} is introduced in vision domain, which designs visual prompts in the input sequence and shows competitive performance with full fine-tuning. However, current methods do not consider the inner design of transformer-based architectures, resulting in less effective prompting solutions. In contrast, our approach is mindful of architecture and anchored on pruning, which conceptually sets it apart from the methods discussed above.

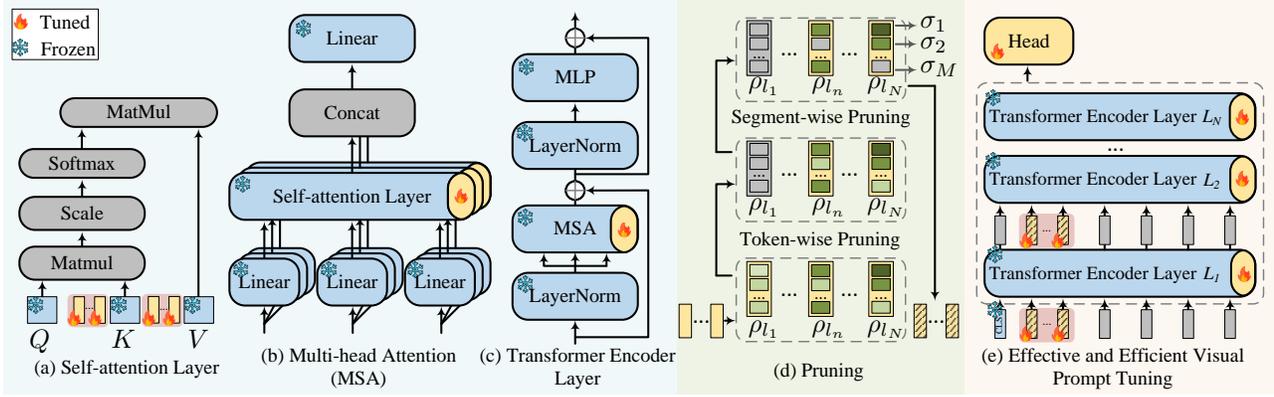
\begin{figure*}[t!]
  \centering
\includegraphics[width=1.0\textwidth]{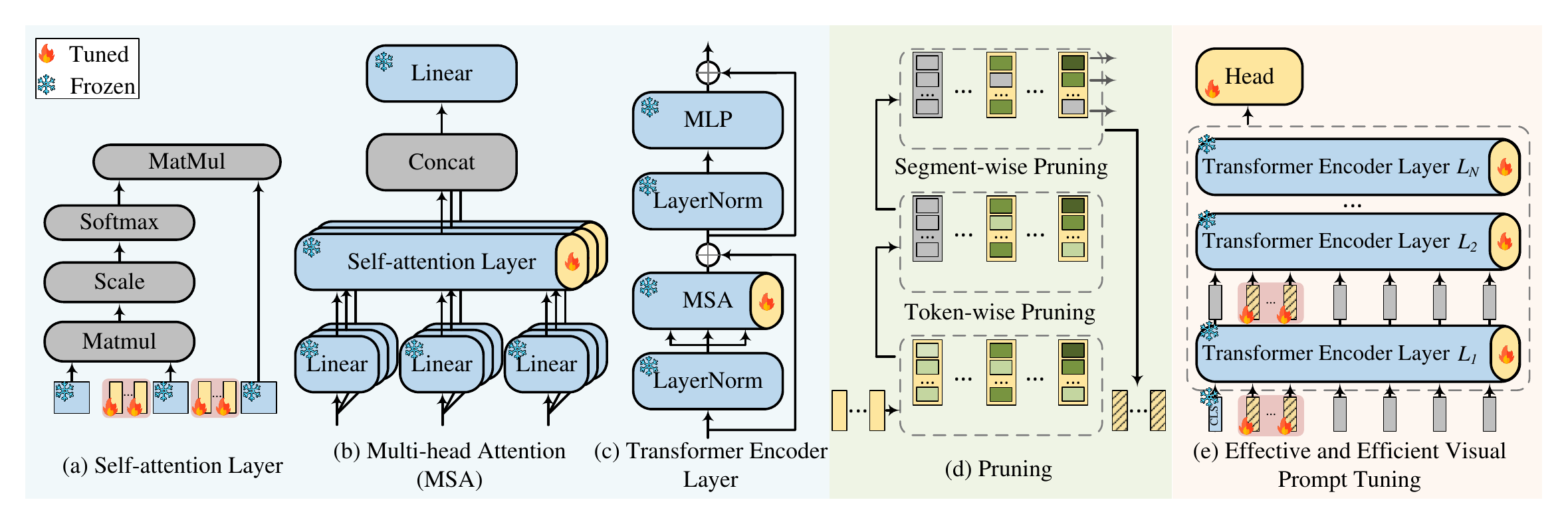}
\put(-207,32){$\rho_{l_{1}}$}
\put(-207,78){$\rho_{l_{1}}$}
\put(-207,124){$\rho_{l_{1}}$}
\put(-183,32){$\rho_{l_{n}}$}
\put(-183,78){$\rho_{l_{n}}$}
\put(-183,124){$\rho_{l_{n}}$}
\put(-162,32){$\rho_{l_{N}}$}
\put(-162,78){$\rho_{l_{N}}$}
\put(-162,124){$\rho_{l_{N}}$}
\put(-142,147){$\sigma_{1}$}
\put(-142,139){$\sigma_{2}$}
\put(-142,130){$\sigma_{M}$}
\put(-479,26){$Q$}
\put(-448,26){$K$}
\put(-419,26){$V$}
\caption{\textbf{Overview of our \textbf{E$^{2}$VPT} framework.} Under the \textit{pretrain-then-finetune} paradigm, only the prompts in the transformer's input and backbone (\S\ref{subsec:effective_prompting}), are updated during the fine-tuning process, while all other components remain frozen. We further introduce pruning (\S\ref{subsec:Pruning_Rewind}) at two levels of granularity (\ie, token-wise and segment-wise) in (d) to eliminate unfavorable input prompts during rewinding.}
\label{fig:plot2}
\vspace{-10pt}
\end{figure*}

\section{Our \textbf{E$^{2}$VPT} Approach}
\label{sec:mixed_prompt}
\vspace{-0.3em}
In this section, we introduce ${\rm E^{2}VPT}$, a novel visual prompt tuning approach for effective and efficient large-scale transformer-based model fine-tuning. We first define the problem and notations in \S\ref{subsec:attention_anatomy}. The effective prompt tuning with the designing of visual and key-value prompts is presented in \S\ref{subsec:effective_prompting}, followed by the efficient prompt pruning in \S\ref{subsec:Pruning_Rewind}. The overall framework is shown in Fig. \ref{fig:plot2}.

\subsection{Problem Definition}
\label{subsec:attention_anatomy}
\vspace{-0.2em}
In this section, we define the problem of ${\rm E^{2}VPT}$ and provide the notations.
Assuming we have a backbone vision transformer model $\textbf{T}$, which is pretrained on a large set of data and tasks. The input to the vision transformer is a sequence of image patches $I$ = $\{I_1, I_2,\dots,I_m\}$, where $m$ is the total number of image patches. Each patch is then projected into a $d$-dimensional embedding with positional encoding, \ie, $E$ = $\{E_j | 1\le j\le m\}$ with $E_j$ = ${\rm Emb}(I_j)$. The vision transformer $\textbf{T}$ consists of $N$ identical transformer layers, represented as:
\begin{equation}
\begin{aligned}
    &Z^1 = {L_1}(E) \\
    &Z^i = {L_i}(Z^{i-1}) \ \ \ \ \ i = 2, 3,\dots,N
\end{aligned}
\end{equation}
here each transformer layer is a stack of multi-head self-attention (\text{MSA}) and feed-forward network (\text{FFN}):
\begin{equation}
L(\cdot) = \text{FFN} \ ( \ \text{MSA}\ (\cdot)\ )
\end{equation}
Given a new vision task, the objective is to fine-tune a model $\hat{\textbf{T}}$ that can deliver good performance on the task, while only tuning a small amount of parameters. 
In the context of visual prompt tuning, $\hat{\textbf{T}}$=$\{\textbf{\textcolor{iceblue}{T}}, \textbf{\textcolor{fired}{P}}\}$ which includes a \textcolor{iceblue}{frozen} backbone \textbf{\textcolor{iceblue}{T}}, and \textcolor{fired}{trainable} prompts \textbf{\textcolor{fired}{P}} with very few tunable parameters.

\subsection{Effective Prompting}
\label{subsec:effective_prompting}
Most existing prompt tuning approaches focus on tuning a set of visual prompts by prepending them to the input sequence in transformer layers, without considering the internal design of transformer architectures. However, to enhance the effectiveness of prompt tuning and achieve optimal fine-tuning performance, we propose a new approach that incorporates a set of key-value prompts (\textbf{{P$_K$}} and \textbf{{P$_V$}}) in addition to input visual prompts (\textbf{{P$_I$}}) within our visual prompt tuning framework.
Intuitively, the input visual prompts are inserted to the input sequence of each encoder layer, which learn to represent of the new task. The key-value prompts are concatenated with the key and value parameter matrices in the self-attention module, which learn to capture the new attention pattern from the data.

\noindent \textbf{Visual Prompts.}
Visual prompts are a set of $d$-dimensional embedding vectors that have the same dimensionality with the input visual tokens. They are prepended to the input sequence of each transformer encoder layer and interact with all the input tokens. Visual prompts play a similar role to those prompt tokens in traditional prompt tuning methods~\cite{jia2022visual,lester2021power}, which learn task-specific embeddings to guide the model performing on the new task.

Formally, these visual prompts are defined as P$_I$ = $\{P_I^1, P_I^2, \dots, P_I^N\}$, where $P_I^i$ denotes the learnable visual prompts in the $i_{th}$ encoder layer, and $N$ is the total number of layers. Then the encoder layers are represented as:
\begin{equation}
\begin{aligned}
    &Z^1 = \textcolor{iceblue}{L_1}(\textcolor{fired}{P_I^1}, \ \textcolor{iceblue}{E}) \\
    &Z^i = \textcolor{iceblue}{L_i}(\textcolor{fired}{P_I^i}, \ Z^{i-1}) \ \ \ \ \ i = 2, 3,\dots,N
\end{aligned}
\end{equation}
where $Z^i$ represents the contextual embeddings computed by the $i_{th}$ encoder layer. The different colors indicate \textcolor{fired}{trainable} and \textcolor{iceblue}{frozen} parameters, respectively. For the embeddings of the input image patches $E$, they are initialized with frozen ${\rm Emb}$ projection from the backbone.
\noindent \textbf{Key-Value Prompts.}
Visual prompts are useful in learning knowledge about new tasks. However, they are insufficient in guiding information interaction within transformer encoder layers. The reason is that when fine-tuning on new data, the image distribution may significantly differ from those in the image examples used for pretraining the backbone model. As a result, it is crucial to enhance the model's capability to capture new information from the fine-tuning data and conduct more effective attention among input tokens to learn new patterns.

To this end, we propose a new approach by introducing a novel set of key-value prompts, P$_K$ and P$_V$, which are incorporated into the attention module within each encoder layer (as shown in Fig. \ref{fig:plot2}(a). These key-value prompts are small matrices that have only a few columns but share the same number of rows as the key and value matrices in the original attention module. To perform the new attention computations, the key and value matrices are concatenated with their corresponding P$_K$ and P$_V$ prompts, respectively. This process is defined as follows:
\begin{equation}
\begin{aligned}
    L(\cdot) &= \textcolor{iceblue}{\text{FFN}} \ (\textcolor{fired}{\text{MSA}} \ (\cdot)) \\
    \text{MSA} (\cdot)  &= {\rm concat}({\rm softmax}(\frac{\textcolor{iceblue}{Q_h}\textcolor{fired}{K_h^{'}}^T}{\sqrt{d}})\textcolor{fired}{V_h^{'}})
\end{aligned}
\end{equation}
where \text{FFN} is the feed-forward network and \text{MSA} is the multi-head attention inside the encoder layer. $h$ represents the $h_{th}$ head. $\textcolor{fired}{K^{'}}$ and $\textcolor{fired}{V^{'}}$ are the new key and value embedding matrices defined as:
\begin{equation}
\label{eq:kv_promt}
\begin{aligned}
    K^{'} = {\rm concat}(\textcolor{iceblue}{K}, \ \textcolor{fired}{P_K}), \ \ \ V^{'} = {\rm concat}(\textcolor{iceblue}{V}, \ \textcolor{fired}{P_V})
\end{aligned}
\end{equation}
where $K$ and $V$ represent the original key and value matrices in the backbone. In this way, the key-value prompts can help guide the model adaptation to the new data. 
In our implementation, we take it a step further by enabling parameter sharing of the P$_K$ and P$_V$ prompts within each transformer layer instead of tuning separate learnable vectors.
Our motivation is twofold: First, our experimental results show that with the shared prompts, the fine-tuning performance consistently improves across instances; Second, using shared prompt vectors reduces the parameter usage in the learnable transformer part by half, making it more parameter-efficient. We provide discussion on exploring the prompt locations (i.e., before or after $K$ and $V$) in \S\ref{subsec:Diagnostic_Experiment}. 

It is worth noting that the query matrix $Q$ is another critical element in the self-attention mechanism. However, additional prompting on $Q$ is not desired for two reasons: First, prompting on $Q$ is similar to prepending on $K$ for computing attention scores between each pair of $Q$ and $K$ Therefore, prompting on both $Q$ and $K$ is unnecessary; Second, changes in $Q$ affect the output shape of the attention map, necessitating an additional linear projection for unmatched dimensions in the following layer. This is not affordable under the parameter-efficient design. More experiments and discussions will be provided in the Appendix.



\subsection{Efficient Prompting}
\label{subsec:Pruning_Rewind}

Our approach to effective prompting aims to enhance the performance of the fine-tuned model. However, a natural question arises: Can we reduce the number of tunable prompts without sacrificing model performance? The lottery ticket hypothesis (LTH)~\cite{frankle2018lottery, zhuang2023survey} states that there exists a sub-network that can achieve the same test performance as the original over-parameterized network for a given task, without the need for unnecessary weights. Motivated by this hypothesis, we conducted an experiment in which we masked different visual prompts and found that various prompts have varying effects on the model performance, with some even having a negative impact. This observation is consistent with previous research~\cite{li2022automated,ma2022xprompt}.

Based on our findings, we propose a prompt pruning method on visual prompts. 
The primary objective of this method is to retain the most influential prompts while eliminating redundant or unnecessary ones. By removing less important prompts, we can significantly improve the efficiency of prompt tuning while maintaining performance.
To achieve this goal, we design a cascade pruning strategy that operates at two levels of granularity, namely token-wise pruning and segment-wise pruning, as illustrated in Fig.~\ref{fig:plot2}(d). Token-wise pruning initially identifies and removes the least important visual prompts. After this step, segment-wise pruning divides each remaining prompt into multiple segments and filters out negative segments. By jointly reducing the parameter usage in learnable visual prompts, our two-level pruning approach creates soft filtered prompts that can be re-trained in the rewinding stage.

\noindent \textbf{Token-wise Pruning.} We introduce a learnable mask variable $\rho=\left\{\rho_{1},\rho_{2},\dots,\rho_{M}\right\}$ ($M$ is the length of visual prompts) and associate it with the input visual prompts in each transformer layer. Here $\rho_{k} \in \left\{0,1\right\}$, where 0 means the corresponding learnable input prompt is pruned. Then the masked version of the visual prompts becomes $\widetilde{P_k}$ = $\rho_{k} \cdot P_k$.
To determine the pruning position, we  
calculate the importance score~\cite{frankle2018lottery, ma2022xprompt} of each prompt token and eliminate those positions with lowest scores.
The importance score is defined as the expected sensitivity of the model to the mask variables $\rho_{k}$ \cite{michel2019sixteen}:
\begin{equation}
\label{eq:pruning_2}
S_{P_k} = \mathbb{E}_{x \sim \mathcal{D}_{x}}\left| \frac{\partial \mathcal{L}(x)}{\partial \rho_{k}}\right|
\end{equation}
where $\mathcal{L}$ is the loss function, and $\mathcal{D}_{x}$ is the training data distribution~\cite{michel2019sixteen}. 
The importance score assigned to each visual prompt reflects its contribution to the fine-tuning performance. A low importance score indicates that the prompt has a minor or even negative contribution to the fine-tuning process. Conversely, a high importance score suggests that the prompt is a meaningful and useful one that significantly contributes to the fine-tuning process.

\noindent \textbf{Segment-wise Pruning.}  
We further investigate the segment-wise pruning to preclude the negative prompt segments within each prompt. The embedding of each prompt token is first equally divided into $R$ parts. Each part is treated as an isolated unit which can be optimized jointly. Similar to the token-wise pruning, we then assign a mask variable to each segment inside the prompt token and filter out those segments with low importance scores.


\noindent \textbf{Rewinding.} 
After performing the two-level cascade pruning, the weight rewinding stage focuses on re-training the soft filtered prompt tokens. This process involves ranking the importance scores for each layer during the pruning stage and setting the corresponding mask variables to 0 when their importance scores are relatively low. Next, the soft filtered input prompts are re-trained along with other learnable parameters using the original combination of learning rate and weight decay during fine-tuning.

\begin{table*}[t]
\caption{\textbf{Image classification accuracy for ViT-Base/16~\cite{dosovitskiy2020image}} pretrained on supervised ImageNet-21k. Following \cite{jia2022visual}, we report the average test accuracy (three runs) on FGVC~\cite{jia2022visual} and VTAB-1k~\cite{zhai2019large} benchmarks, and ``Number of Wins" in [$\cdot$] compared to full fine-tuning (Full)~\cite{iofinova2022well}. ``Tuned/Total" is the average percentage of tuned parameters required by 24 tasks. ``Scope” indicates the tuning scope of each method. ``Additional parameters" is the existence of parameters in addition to the pretrained backbone and linear head. The highest accuracy among all approaches except FULL are shown in \textbf{bold}. ${\rm E^{2}VPT}$ outperforms the full fine-tuning in \textbf{19 of 24} instances with far fewer trainable parameters. More impressively, we further report ``Number of Wins to VPT" in $\left\{\cdot\right\}$. Our method beats VPT in \textbf{21 of 24} cases with considerably lower parameters. Per-task results are available in Appendix. Same for Table \ref{table:swin} and \ref{table:mae_moco}.}
\vspace{5pt}
\label{table:fgvc_vtab_main}
\begin{adjustbox}{width=0.83\width,center}
\begin{tabular}{c||r|cc|c|r|rrr} 
\hline \thickhline
\rowcolor{mygray}
ViT-Base/16~\cite{dosovitskiy2020image}     & Tuned/ &\multicolumn{2}{c|}{Scope}   & Extra    &   &  \multicolumn{3}{c}{VTAB-1k~\cite{zhai2019large} [19]}  \\ 
\rowcolor{mygray}
\rowcolor{mygray}
  (85.8M)   & Total & Input & Backbone  & params  &  \multirow{-2}{*}{FGVC~\cite{jia2022visual} [5]}    &  \textit{Natural} [7] & \textit{Specialized} [4] & \textit{Structured} [8]   \\ 
\hline \hline
Full \textcolor{lightgray}{\scriptsize{[CVPR22]}}\cite{iofinova2022well} & 100.00\% & & \checkmark & & 88.54\% & 75.88\% & 83.36\% & 47.64\% \\
\hline
Linear \textcolor{lightgray}{\scriptsize{[CVPR22]}}\cite{iofinova2022well} & 0.08\% & & & & 79.32\% [0] & 68.93\% [1] & 77.16\% [1] & 26.84\% [0] \\
Partial-1 \textcolor{lightgray}{\scriptsize{[NeurIPS14]}}\cite{yosinski2014transferable} & 8.34\% & & & & 82.63\% [0] & 69.44\% [2] & 78.53\% [0] & 34.17\% [0] \\
MLP-3 \textcolor{lightgray}{\scriptsize{[CVPR20]}}\cite{chen2020improved} & 1.44\% & & & \checkmark & 79.80\% [0] & 67.80\% [2] & 72.83\% [0] & 30.62\% [0]\\
\hline
Sidetune \textcolor{lightgray}{\scriptsize{[ECCV20]}}\cite{zhang2020side} & 10.08\% & & \checkmark & \checkmark & 78.35\% [0] & 58.21\% [0] & 68.12\% [0] & 23.41\% [0] \\
Bias \textcolor{lightgray}{\scriptsize{[NeurIPS17]}}\cite{rebuffi2017learning} & 0.80\% & & \checkmark & & 88.41\% [3] & 73.30\% [3] & 78.25\% [0] & 44.09\% [2]\\
Adapter \textcolor{lightgray}{\scriptsize{[NeurIPS20]}}\cite{cai2020tinytl} & 1.02\% & & \checkmark & \checkmark & 85.66\% [2] & 70.39\% [4] & 77.11\% [0] & 33.43\% [0]\\
\hline
VPT \textcolor{lightgray}{\scriptsize{[ECCV22]}}\cite{jia2022visual} & 0.73\% & \checkmark &  & \checkmark & 89.11\% [4] & 78.48\% [6] & 82.43\% [2] & 54.98\% [8]\\
\rowcolor{mygray}
Ours & 0.39\% & \checkmark & \checkmark & \checkmark & \textbf{89.22\%} [4] $\left\{4\right\}$ & \textbf{80.01\%} [6] $\left\{5\right\}$ & \textbf{84.43\%} [3] $\left\{4\right\}$ & \textbf{57.39\%} [8] $\left\{7\right\}$ \\
\hline
\end{tabular}
\end{adjustbox}
\vspace{-1em}
\end{table*}

\begin{table}[t]
\caption{\textbf{Image classification accuracy for Swin-Base~\cite{liu2021swin}} pretrained  on supervised ImageNet-21k. }
\vspace{-15pt}
\label{table:swin}
\begin{center}
\begin{small}
\tabcolsep=0.10cm
\resizebox{0.49\textwidth}{!}{
\begin{tabular}{c||r|rrr} 
\hline \thickhline
\rowcolor{mygray}
Swin-Base~\cite{liu2021swin}  & Tuned/ &  \multicolumn{3}{c}{VTAB-1k~\cite{zhai2019large} [19]}  \\ 
\rowcolor{mygray}
\rowcolor{mygray}
  (86.7M)   & Total &  \textit{Natural} [7] & \textit{Specialized} [4] & \textit{Structured} [8]   \\ 
\hline \hline
Full \textcolor{lightgray}{\scriptsize{[ICLR23]}}\cite{ren2023prepare} & 100.00\% &  79.10\% & 86.21\% & 59.65\% \\
\hline
Linear \textcolor{lightgray}{\scriptsize{[ICLR23]}}\cite{ren2023prepare} & 0.06\% & 73.52\% [5] & 80.77\% [0] & 33.52\% [0]  \\
Partial-1 \textcolor{lightgray}{\scriptsize{[NeurIPS14]}}\cite{yosinski2014transferable} & 14.58\% & 73.11\% [4] & 81.70\% [0] & 34.96\% [0]  \\
MLP-3 \textcolor{lightgray}{\scriptsize{[CVPR20]}}\cite{chen2020improved} & 2.42\% &  73.56\% [5] & 75.21\% [0] & 35.69\% [0] \\
\hline
Bias \textcolor{lightgray}{\scriptsize{[NeurIPS17]}}\cite{rebuffi2017learning} & 0.29\% &  74.19\% [2] & 80.14\% [0] & 42.42\% [0]  \\
\hline
VPT \textcolor{lightgray}{\scriptsize{[ECCV22]}}\cite{jia2022visual} & 0.25\% & 76.78\% [6] & 83.33\% [0] & 51.85\% [0]  \\
\rowcolor{mygray}
Ours &  0.21\% &  \textbf{83.31\%} [6] $\left\{6\right\}$ & \textbf{84.95\%} [2] $\left\{3\right\}$ & \textbf{57.35\%} [3] $\left\{7\right\}$ \\
\hline
\end{tabular}
}
\end{small}
\end{center}
\vspace{-1.5em}
\end{table}

\begin{table*}[t]
\caption{\textbf{Image Classification accuracy for different pretrained objectives} --- MAE~\cite{he2022masked} and MoCo v3~\cite{chen2021empirical} with ViT-Base~\cite{dosovitskiy2020image} as backbone. Our method enjoys significant performance gains to VPT~\cite{jia2022visual} while having lower parameter usage.}
\vspace{-15pt}
\label{table:mae_moco}
\begin{center}
\begin{small}
\tabcolsep=0.10cm
\resizebox{1.0\textwidth}{!}{
\begin{tabular}{c||r|rrr||r|rrr} 
\hline \thickhline
\rowcolor{mygray}
Pretrained objectives & \multicolumn{4}{c||}{MAE~\cite{he2022masked}} &  \multicolumn{4}{c}{MoCo v3~\cite{chen2021empirical}} \\
\hline 
\rowcolor{mygray}
& Tuned/ &  \multicolumn{3}{c||}{VTAB-1k~\cite{zhai2019large} [19]} & Tuned/ &  \multicolumn{3}{c}{VTAB-1k~\cite{zhai2019large} [19]}  \\ 
\rowcolor{mygray}
\rowcolor{mygray}
  \multirow{-2}{*}{\diagbox{Methods~}{Parms \& Data}} & Total &  \textit{Natural} [7] & \textit{Specialized} [4] & \textit{Structured} [8] & Total &  Natural [7] & Specialized [4] & Structured [8]  \\ 
\hline \hline
Full \textcolor{lightgray}{\scriptsize{[CVPR22]}}\cite{iofinova2022well} & 100.00\% &  59.31\% & 79.68\% & 53.82\% & 100.00\% & 71.95\% & 84.72\% & 51.98\% \\
\hline
Linear \textcolor{lightgray}{\scriptsize{[CVPR22]}}\cite{iofinova2022well} & 0.04\% &  18.87\% [0] & 53.72\% [0] & 23.70\% [0] & 0.04\% &  67.46\% [4] & 81.08\% [0] & 30.33\% [0] \\
Partial-1 \textcolor{lightgray}{\scriptsize{[NeurIPS14]}}\cite{yosinski2014transferable} & 8.30\% & 58.44\% [5] & \textbf{78.28\%} [1] & 47.64\% [1] & 8.30\% &  72.31\% [5] & 84.58\% [2] & 47.89\% [1] \\
\hline
Bias \textcolor{lightgray}{\scriptsize{[NeurIPS17]}}\cite{rebuffi2017learning} & 0.16\% &  54.55\% [1] & 75.68\% [1] & \textbf{47.70\%} [0] & 0.16\% & 72.89\% [3] & 81.14\% [0] & 53.43\% [4] \\
Adapter \textcolor{lightgray}{\scriptsize{[NeurIPS20]}}\cite{cai2020tinytl} & 0.87\% &  54.90\% [3] & 75.19\% [1] & 38.98\% [0] & 1.12\% & 74.19\% [4] & 82.66\% [1] & 47.69\% [2] \\
\hline
VPT \textcolor{lightgray}{\scriptsize{[ECCV22]}}\cite{jia2022visual} & 0.10\% &  36.02\% [0] & 60.61\% [1] & 26.57\% [0] & 0.06\% &  70.27\% [4] & 83.04\% [0] & 42.38\% [0]\\
\rowcolor{mygray}
Ours &  0.07\% & \textbf{59.52\% [4]} $\left\{6\right\}$ & 77.80\% [1] $\left\{2\right\}$ & 44.65\% [3] $\left\{8\right\}$ & 0.13\% & \textbf{76.47\%} [4] $\left\{7\right\}$ & \textbf{87.28\%} [2] $\left\{4\right\}$  & \textbf{54.91\%} [6] $\left\{8\right\}$ \\
\hline
\end{tabular}
}
\end{small}
\end{center}
\vspace{-1.5em}
\end{table*}


\section{Experiment}
\label{sec:Experiment}


\subsection{Experimental Setup}
\label{subsec:Experimental_Setup}

\noindent \textbf{Datasets}.$_{\!}$ Our$_{\!}$ experiments$_{\!}$ are$_{\!}$ carried$_{\!}$ out$_{\!}$ on$_{\!}$ two$_{\!}$ image classification benchmarks. 
\textbf{VTAB-1k}~\cite{zhai2019large} collects 19 benchmarked Visual Task Adaptation, categorized into three groups: (1) \textit{Natural} contains natural images captured by standard cameras, (2) \textit{Specialized} includes images taken by specialized equipment, and (3) \textit{Structured} covers tasks requiring geometric comprehension (\ie, counting, distance). Each task of VTAB-1k contains 1000 training examples. Following~\cite{jia2022visual, zhai2019large}, we apply the 800-200 split for training set on hyperparameter tuning. The final evaluation is run on the full training data.
\textbf{FGVC} contains 5 benchmarked Fine-Grained Visual Classification, including CUB-200-2011~\cite{wah2011caltech}, NABirds~\cite{van2015building}, Oxford Flowers~\cite{nilsback2008automated}, Stanford Dogs~\cite{khosla2011novel} and Stanford Cars~\cite{gebru2017fine}. Following~\cite{jia2022visual}, the training set is randomly split into 90\% \texttt{train} and 10\% \texttt{val}. We use \texttt{val} for hyperparameter tuning.


\noindent \textbf{Baselines.}
For fair comparison, we follow \cite{jia2022visual} and compare ${\rm E^{2}VPT}$ with other widely applied parameter-efficient fine-tuning methods. Results of two vision transformer architectures, Vision transformer~\cite{dosovitskiy2020image} (ViT) and Swin transformer~\cite{liu2021swin} (Swin), on image classification are discussed in \S\ref{subsec:Comparison_SOTA}. We also apply ${\rm E^{2}VPT}$ to two self-supervised objectives: MAE~\cite{he2022masked} and MoCo v3~\cite{chen2021empirical}. 

\noindent \textbf{Training.} 
Following \cite{jia2022visual, mahajan2018exploring}, we conduct grid search to match the best tuning hyperparameters, learning rate (\ie, [50, 25, 10, 5, 2.5, 1, 0.5, 0.25, 0.1, 0.05]), and weight decay (\ie, [0.01, 0.001, 0.0001, 0.0]) on \texttt{val} set for each task. Notably, ${\rm E^{2}VPT}$ \textbf{does not require} specific-designed large learning rate in \cite{jia2022visual}. For all models, the learning rate is scheduled following a cosine decay policy and trained for 100 epochs (including 10 warm-up epochs). We follow the same batch size setting in \cite{jia2022visual}: 64/128 for ViT-Base/16 and 80 for Swin-Base, respectively. The number of segments for each token (\S\ref{subsec:Pruning_Rewind}) is set to 8. The percentages for the pruning stage are searched linearly between 10\% and 90\% with 10\% intervals. The rewinding stage applies once to re-train the pruned input prompts. 

\noindent \textbf{Reproducibility.} ${\rm E^{2}VPT}$ is implemented in Pytorch~\cite{NEURIPS2019_9015}. Experiments are conducted on NVIDIA A100-40GB GPUs. To guarantee reproducibility, our full implementation will be publicly released.

\subsection{Comparison with State-of-the-Arts}
\label{subsec:Comparison_SOTA} 
We respectively examine the performance and robustness of ${\rm E^{2}VPT}$ on ViT~\cite{dosovitskiy2020image}, Swin~\cite{liu2021swin}, and two self-supervised objectives --- MAE~\cite{he2022masked} and MoCo v3~\cite{chen2021empirical}. For reference, we provide the individual per-task results for Table~\ref{table:fgvc_vtab_main}, \ref{table:swin} and \ref{table:mae_moco} in Appendix.

\noindent \textbf{\textbf{E$^{2}$VPT} on ViT.} We report the average accuracy score on VTAB-1k and FGVC benchmarks across four diverse task groups for three runs in Table~\ref{table:fgvc_vtab_main}, considering ${\rm E^{2}VPT}$ to the other eight tuning protocols under \textit{pretrain-then-finetune} paradigm. Specifically, Full~\cite{iofinova2022well} updates both backbone and classification head; Linear~\cite{iofinova2022well}, Parital-$1$~\cite{yosinski2014transferable} (top layer) and MLP-$3$~\cite{chen2020improved} (3 MLP layers) are partial tuning methods that only update partial parameters. Sidetune~\cite{zhang2020side}, Bias~\cite{rebuffi2017learning} and Adapter~\cite{cai2020tinytl} are extra module methods which add new trainable parameters to backbone for adaptation; VPT~\cite{jia2022visual} is a most recent visual prompt tuning method.
There are several key observations from these results. \textbf{First}, ${\rm E^{2}VPT}$ is able to outperform the full fine-tuning method in most cases, 21 out of 24 tasks. For example, our model achieves \textbf{0.68\%} improvement on FGVC and \textbf{9.75\%} improvements on VTAB-1k Structured respectively. This observation demonstrates the effectiveness of our approach for fast large-scale vision model adaptation. On the other hand, our model only trains \textbf{0.39\%} parameters in the backbone, which is much more parameter efficient than the full fine-tuned model. \textbf{Second}, it is not surprising to see that the prompt tuning based approaches generally outperform the other parameter efficient methods, such as partial fine-tuning (Partial-1) and extra module (Adapter), indicating the superior adaptability of prompt tuning methods on large-scale vision models. Again, the number of tunable parameters in prompt tuning methods is also smaller compared to the other methods. \textbf{Third}, our approach consistently outperforms the strong VPT model with less tunable prompts, demonstrating the effective design of the key-value prompting and the efficient prompt pruning. The reason is that VPT only focus on design input visual prompts, which fail to capture the accurate interactions between image patches in the new data. In contrast, the key-value prompts in ${\rm E^{2}VPT}$ effectively bridge this gap.

\begin{table*}[t]
\caption{\textbf{Impact of different components} in ${\rm E^{2}VPT}$ on two instances: VTAB-1k \textit{Natural} SVHN~\cite{netzer2011reading} and FGVC NABirds~\cite{nabirds}.}
\label{table:ablative_components}
\begin{center}
\begin{small}
\tabcolsep=0.20cm
\resizebox{0.96\textwidth}{!}{
\begin{tabular}{c|c|c||r|r|r||r|r|r} 
\hline \thickhline
\rowcolor{mygray}
\multicolumn{3}{c||}{Fine-tuning Techniques}  & \multicolumn{3}{c||}{VTAB-1k \textit{Natural} SVHN~\cite{netzer2011reading}} & \multicolumn{3}{c}{FGVC NABirds~\cite{nabirds}}\\ 
\hline 
\rowcolor{mygray}
\rowcolor{mygray}
Visual Prompts  & Key-Value Prompts & Pruning \& Rewinding & Pruning & Tuned / Total  & Accuracy &  Pruning & Tuned / Total  & Accuracy  \\
\hline \hline
\checkmark & & & 0.0\% & 0.54\% & 78.1\% & 0.0\% & 1.02\% & 84.2\% \\
\checkmark & \checkmark & & 0.0\% & 0.55\% & 83.8\% & 0.0\% & 1.05\% & 84.5\% \\
\checkmark & & \checkmark & 56.3\% & 0.42\% & 79.0\% & 34.4\% & 0.63\% & 84.2\%\\
\checkmark & \checkmark & \checkmark & 62.5\% & 0.43\% & \textbf{85.3\%} & 40.0\% & 0.65\%  & \textbf{84.6\%} \\
\hline
\end{tabular}
}
\end{small}
\end{center}
\vspace{-2.5em}
\end{table*}



\noindent \textbf{\textbf{E$^{2}$VPT} on Hierarchical Transformer.} To prove the effectiveness and generalization of our architectural design, we further extend ${\rm E^{2}VPT}$ to a hierarchical transformer --- Swin~\cite{liu2021swin}, where the MSA layer is employed in local shifted windows and patch embeddings are merged at deeper layers. For generality, we follow the same settings in ViT~\cite{dosovitskiy2020image} architecture to prepend K-V learnable pairs and \cite{jia2022visual} for altering input vectors (\ie, these learnable vectors are attended within the local windows and ignored during patch merging). For pruning, we notice performance drop when incorporating within the deeper local windows. We therefore assign pruning stage only to the first stage. As Swin does not use \texttt{[CLS]} and apply the global pooling as input for classification head~\cite{jia2022visual, liu2021swin}, we follow this design when adapting our method. The exclusive experiments are deployed on the ImageNet-21k supervised pretrained Swin-Base~\cite{liu2021swin}. ${\rm E^{2}VPT}$ consistently outperform all the other parameter-efficient methods on all three VTAB-1k problem classes and for the first time surpasses full fine-tuning on VTAB-1k \textit{Specialized} and \textit{Structured} using significantly fewer parameters (\ie, \textbf{0.21\%}).


\noindent \textbf{Different Pretraining Methods.}
We conducted experiments with two self-supervised objectives, MAE~\cite{he2022masked} and MoCo v3~\cite{chen2021empirical}, on backbones pretrained without labeled data, following the approach of VPT~\cite{jia2022visual}. While VPT yielded inconclusive results on these objectives, our proposed method, ${\rm E^{2}VPT}$, outperformed other methods and achieved competitive performance to full fine-tuning (\textbf{8 of 19} instances under MAE, and \textbf{12 of 19} instances under MoCo v3), using significantly fewer model parameters (\textbf{0.07\%} on MAE and \textbf{0.13\%} on MoCo v3). Our method also outperformed VPT by a large margin (\textbf{59.52\%} vs. 36.02\% under MAE on VTAB-1k \textit{Natural}). We leveraged the gap discussed in VPT, which indicates that self-supervised ViTs are fundamentally different from the supervised ones, and demonstrated the generality of our method to both pretraining objectives.

\subsection{Diagnostic Experiments}
\label{subsec:Diagnostic_Experiment}



\noindent \textbf{Impact of Different Components.} 
To investigate the impact of different components in ${\rm E^{2}VPT}$, including visual prompts, key-value prompts, and pruning and rewinding, we conducted experiments on two tasks in the benchmarks. The results are summarized in Table~\ref{table:ablative_components}. For SVHN~\cite{netzer2011reading}, we found that the model with visual prompts alone achieved an accuracy of 78.1\%. Adding key-value prompts and applying pruning and rewinding techniques individually led to additional gains (\textbf{5.7\%} and \textbf{0.9\%}), demonstrating the effectiveness of our key-value prompt tuning technique in the self-attention module as well as the pruning mechanism. Finally, combining all components together yielded the best performance, with an accuracy of 85.3\%. We observed similar trends on FGVC NABirds~\cite{nabirds}.

\noindent \textbf{Prompt Location.} An fundamental distinction between ${\rm E^{2}VPT}$ and other methods is the learnable key-value prompts introduced to self-attention. In our implementation, we prepend the key-value prompts to the sequence of Key and Value matrices. Further investigation is required to determine the appropriate placement of the learnable prompts. We provide ablation results on VTAB-1k exhaustively in Table~\ref{table:before_after}(a). We show that both prepending learnable prompts before or after Key and Value matrices show competitive results, validating the robustness of our approach on prompt locations. We choose ``Before" as our baseline method in all our experiments since it achieves slightly better results on average (\ie, 73.94\% $vs$ 73.91\%).

\begin{table}[t]
\caption{\textbf{Prompt location and Initialization} results on VTAB-1k~\cite{zhai2019large} in three runs. Per-task results are available in Appendix.}
\vspace{-15pt}
\label{table:before_after}
\begin{center}
\begin{small}
\tabcolsep=0.10cm
\resizebox{0.49\textwidth}{!}{
\begin{tabular}{rr||ccc} 
\hline \thickhline
\rowcolor{mygray}
\multicolumn{2}{c||}{ViT-Base/16~\cite{dosovitskiy2020image}}   &  \multicolumn{3}{c}{VTAB-1k~\cite{zhai2019large} [19]}  \\ 
\rowcolor{mygray}
\rowcolor{mygray}
  \multicolumn{2}{c||}{(85.8M)}     &  \textit{Natural} [7] & \textit{Specialized} [4] & \textit{Structured} [8]   \\ 
\hline \hline
 & After & \textbf{80.67\%} [6] & 84.30\% [3] & 56.76\% [8] \\
\multirow{-2}{*}{(a)}& Before &  80.01\% [6] & \textbf{84.43\%} [3]  & \textbf{57.39\%} [8] \\
\hline
 & \textit{Trunc. Norm.}~\cite{NEURIPS2019_9015} &  79.77\% [6]  &  84.30\% [3] & 56.36\% [8] \\
\multirow{-2}{*}{(b)}& \textit{He}~\cite{he2015delving} &   \textbf{80.01\%} [6] & \textbf{84.43\%} [3] & \textbf{57.39\%} [8]  \\
\hline
\end{tabular}
}
\end{small}
\end{center}
\vspace{-2.5em}
\end{table}

\begin{figure*}[t]
  \centering
\includegraphics[width=0.99\textwidth]{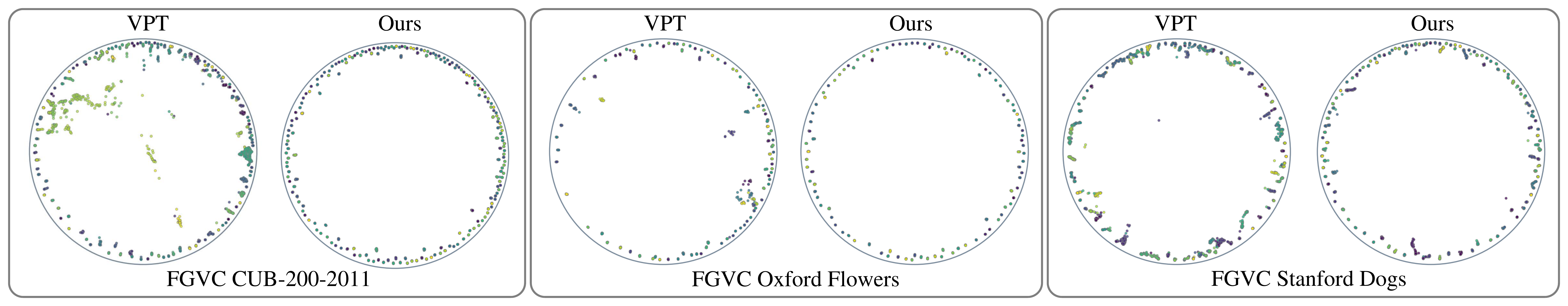}
\caption{\textbf{Hyperbolic visualization results} from VPT~\cite{jia2022visual} and ours on 3 FGVC tasks (\ie, FGVC CUB-200-2011~\cite{wah2011caltech}, Oxford Flowers~\cite{nilsback2008automated} and Stanford Dogs~\cite{khosla2011novel}). Our method shows consistently better clustering pushed to the border of the Poincar\'e disk.}
\label{fig:visulization}
\vspace{-15pt}
\end{figure*}
\begin{figure}
  \centering
       \vspace{-10pt}
\includegraphics[width=0.4\textwidth]{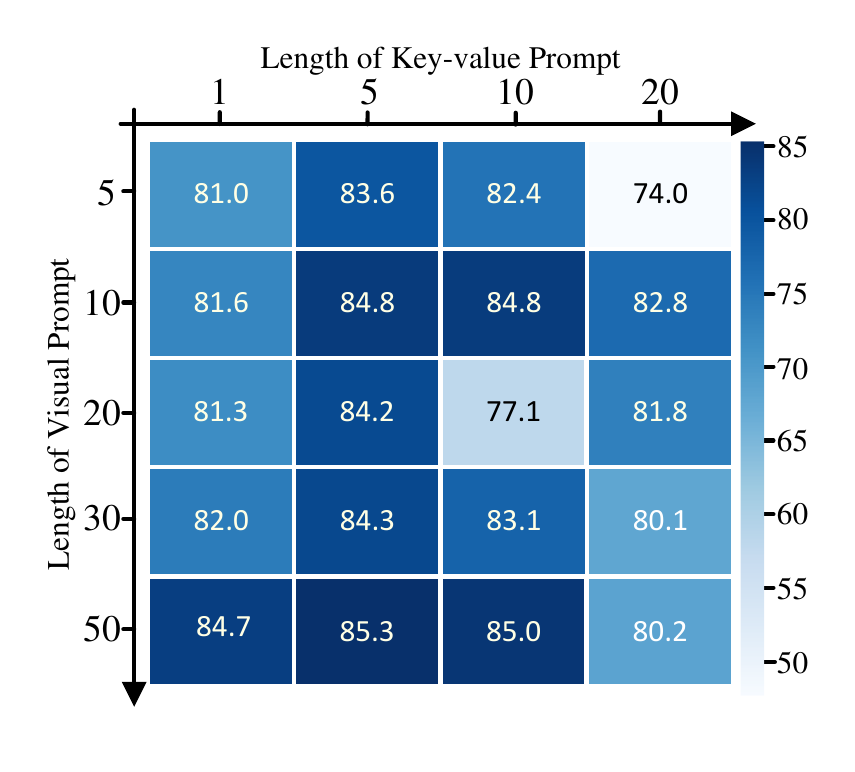}
\vspace{-10pt}
\caption{\textbf{Sensitivity of input prompt and key-value prompt lengths.} We vary the number of prompts for different combinations, and show their results on VTAB-1k $Natural$ SVHN~\cite{netzer2011reading}.}
\label{fig:prompt_length}
\vspace{-15pt}
\end{figure}

\noindent \textbf{Initialization.} Table~\ref{table:before_after}(b) reports the performance of our approach with respect to two widely adopted initialization methods: 
  \textit{truncated normal}~\cite{narkhede2022review, NEURIPS2019_9015} and \textit{He initialization}~\cite{he2015delving} on VTAB-1k benchmark. The results show that \textit{He initialization} generally provides more stable and preferable performances on average, though we observe that in some specific tasks (\ie, \textit{truncated normal} is 1.1\% higher in accuracy over \textit{He} on VTAB-1k \textit{Specialized} Diabetic Retinopathy Detection~\cite{diabetic-retinopathy-detection}) \textit{truncated normal} gets slightly better results. In conclusion, ${\rm E^{2}VPT}$ shows robustness on different initialization methods and is able to achieve consistent performance with full fine-tuning.

\noindent \textbf{Prompt Length.} 
Prompt length is the only hyper-parameter needed to tune in ${\rm E^{2}VPT}$. To further analyze the impact of different prompt lengths on the model performance, we conducted a comprehensive study on the lengths of visual prompts and key-value prompts for a better understanding of their characteristics on VTAB-1k \textit{Natural} SVHN~\cite{netzer2011reading}. The length of visual prompts is typically limited to [5, 10, 20, 30, 50], while the length of key-value prompts is restricted to [1, 5, 10, 50], which is a standard configuration for most datasets. The model performance results on different prompt length combinations are reported in Fig.~\ref{fig:prompt_length}. It can be seen that, when using 50 visual prompts, a relative shorter key-value prompt can benefit performance notably (\ie, 84.7\% when introducing one key-value prompt $vs$ 78.1\% without key-value prompts), while further increasing the length of the key-value prompt yields a small performance gain (\ie, 85.3\% when using 5 key-value prompts). We also notice that using a large number of key-value prompts lead to subpar results (\ie, 80.2\% with 20 key-value prompts). Similar patterns are observed with other visual prompt lengths. 
We argue that a heavy parameter engineering in self-attention layer might distort the original attention map and does harm to adaptation.

\subsection{Visualization}
\label{subsec:Hyperbolic}
Following \cite{atigh2022hyperbolic, ermolov2022hyperbolic, ganea2018hyperbolic, khrulkov2020hyperbolic, peng2021hyperbolic}, we show hyperbolic visualizations results on training set for VPT and ours on three tasks in FGVC (\ie, CUB-200-2011~\cite{wah2011caltech}, Oxford Flowers~\cite{nilsback2008automated}, and Stanford Dogs~\cite{khosla2011novel}). Hyperbolic space, to be specific, is a Riemannian manifold of constant negative curvature. While there are several isometric models of hyperbolic space, we follow previous work~\cite{ermolov2022hyperbolic, ganea2018hyperbolic} and stick to the Poincar\'e ball model.
Similar to~\cite{ermolov2022hyperbolic}, we use UMAP~\cite{mcinnes2018umap} with the ``hyperboloid" distance metric to reduce the dimensionality to 2D. ViT-Base plays as an encoder with two types of pretraining (\ie, tuned models under VPT, and ours after rewinding, respectively). We freeze the models during fine-tuning and output embeddings are mapped to hyperbolic space. Adam optimizer~\cite{loshchilov2017decoupled} with a learning rate of $3 \times 10^{-5}$ is applied to all settings. The weight decay is 0.01 with batch size equals to 900. All models are trained for 50 steps for fair comparison, with a gradient clip by norm 3.

Fig.~\ref{fig:visulization} illustrates how learned embeddings are arranged on the Poincar\'e disk. 
We can see that in ${\rm E^{2}VPT}$, samples are clustered according to labels, and each cluster is pushed closer to the border of the disk, indicating that the encoder separates class well. On the other hand, we observe in VPT that some of the samples move towards the center and intermix~\cite{ermolov2022hyperbolic}, indicating possible confusion during projection. We also follow~\cite{ermolov2022hyperbolic, peng2021hyperbolic, khrulkov2020hyperbolic} and present the Recall@K metric in Appendix for reference. These visualization results further validate the effectiveness of the proposed ${\rm E^{2}VPT}$ approach in generating separatable embeddings from the input images in the new tasks.

\vspace{-1em}
\section{Conclusion and Discussion} \label{sec:conclusion_discussion}
\vspace{-0.3em}
The vast majority of current efforts under the \textit{pretrain-then-finetune} paradigm seek to reduce parameter usage while overlooking the inner design of transformer-based architecture. In light of this view, 
we present ${\rm E^{2}VPT}$, a new parameter-efficient visual prompt tuning approach to model the transformer architecture during adaptation. It enjoys several advantages: \textbf{i)} consider self-attention mechanism during tuning for superior performance to current parameter-efficient fine-tuning; and \textbf{ii)} apply pruning and rewinding stages to reduce parameter usage in input visual prompts. The systemic merits enable an  effective yet efficient algorithm. As a whole, we conclude that the outcomes elucidated in this paper impart essential understandings and necessitate further exploration within this realm.

\noindent \textbf{Acknowledgements.} 
This research was supported by the National Science Foundation under Grant No. 2242243.

{\small
\bibliographystyle{ieee_fullname}
\bibliography{egbib}
}

\end{document}